%% file: main.tex
\documentclass[11pt]{article}

\usepackage[final]{acl}

\usepackage{times}
\usepackage{latexsym}
\usepackage{amsmath}

\usepackage{graphicx}
\usepackage{xcolor}
\PassOptionsToPackage{table}{xcolor}
\usepackage{colortbl}
\usepackage{booktabs}
\usepackage{multirow}
\usepackage{array}
\usepackage{amsmath}
\usepackage{arydshln}
\usepackage{xcolor}
\usepackage{hyperref}
\usepackage{url}
\usepackage{enumitem}
\usepackage{algorithm}
\usepackage{algorithmicx}
\usepackage{algpseudocode}
\usepackage{float}
\usepackage{xspace}

\usepackage{tcolorbox}       
\tcbuselibrary{breakable}    
\usepackage{setspace}        
\usepackage{xcolor}          

\definecolor{bgcolor}{RGB}{242,243,245}

\newcolumntype{B}{>{\columncolor{blue!4}}l}
\newcolumntype{d}{>{\columncolor{brown!4}}l}
\newcolumntype{R}{>{\columncolor{red!4}}l}

\usepackage[T1]{fontenc}

\usepackage[utf8]{inputenc}

\usepackage{microtype}

\usepackage{inconsolata}

\usepackage{amsmath}
\usepackage{amssymb}   

\usepackage{graphicx}

\title{ADaPT: Token-Level Decoupling for Efficient Large Reasoning Models}

\author{
    Tingyun Li\textsuperscript{\rm 1}\footnotemark[1], Zishang Jiang\textsuperscript{\rm 1}\footnotemark[1], Jinyi Han\textsuperscript{\rm 2}, Xinyi Wang\textsuperscript{\rm 1}, Sihang Jiang\textsuperscript{\rm 3}, Han Xia\textsuperscript{\rm 4}, \\ \textbf{ Zhaoqian Dai\textsuperscript{\rm 4}, Shuguang Ma\textsuperscript{\rm 4}, Fei Yu\textsuperscript{\rm 4}, Jiaqing Liang\textsuperscript{\rm 1}\footnotemark[2], Yanghua Xiao\textsuperscript{\rm 3}} \\
    \textsuperscript{\rm 1}School of Data Science, Fudan University\\
    \textsuperscript{\rm 2}Shanghai Institute of Artificial Intelligence for Education, East China Normal University\\
    \textsuperscript{\rm 3}College of Computer Science and Artificial Intelligence, Fudan University \\
    \textsuperscript{\rm 4}Ant Group 
}

\begin{document}
\maketitle
\renewcommand{\thefootnote}{\fnsymbol{footnote}}
\footnotetext[1]{Equal contribution.}
\footnotetext[2]{Corresponding author.}
\renewcommand{\thefootnote}{\arabic{footnote}}
\begin{abstract}


Large reasoning models rely on long chain-of-thought to achieve strong performance, but applying such reasoning uniformly incurs high computational cost. Existing efficiency-oriented methods attempt to shorten or mix reasoning strategies, yet often degrade reasoning capability. We identify the root cause as sequence-level coupling between efficiency incentives and correctness optimization, which implicitly penalizes long but correct reasoning trajectories. To address this issue, we propose Adaptive Dual-Process Thinking (ADaPT), a token-level dual-process framework that explicitly decouples efficiency and correctness signals during training. ADaPT introduces a mode-selection token to control fast and slow reasoning, applying efficiency-related rewards exclusively to this token to avoid penalizing correct long reasoning while encouraging efficiency when appropriate. Moreover, ADaPT enables precise and continuous control over the efficiency–performance trade-off at inference time: by adjusting the generation probability of the mode-selection token, a single trained model can smoothly move along the efficiency–performance Pareto frontier.
Extensive experiments demonstrate that ADaPT significantly reduces inference cost while maintaining strong reasoning performance across multiple benchmarks. Our code is available at \url{https://github.com/SpongeBob-0715/ADaPT_Token_Level_Decoupling}.

\end{abstract}

\section{Introduction}

\input{introduction5}

\input{Preliminary_v2}

\input{method_v7}

\input{experiment3}

\input{relatedwork}

\input{conclusion}
\input{Limitation}
\bibliography{ref}

\appendix

\section*{Appendix}

\input{appendix}

\end{document}

%% file: introduction5.tex

As the focus of language model scaling shifts from training time to test time~\citep{Chen2025TowardsRE,Zhang2025ASO},
large reasoning models (LRMs) have emerged as a distinct class designed for complex reasoning tasks.
Representative examples include OpenAI-o1~\citep{ElKishky2024OpenAIOS} and DeepSeek-R1~\citep{guo2025deepseek}.
These models achieve strong performance primarily by relying on long and structured chain-of-thought reasoning at inference time.

Despite these successes, current large reasoning models suffer from a fundamental inefficiency: they apply long chain-of-thought (CoT) reasoning uniformly across inputs, incurring substantial computational overhead~\citep{Liu2025ThoughtME,Hashemi2025DNRBB,Han2025YourMH}.
For simple queries, such excessive reasoning not only increases token usage and inference cost, but can also harm accuracy due to overthinking~\citep{Chen2024DoNT,Luo2025O1PrunerLF}.
As a result, improving reasoning efficiency without degrading reasoning performance remains a central challenge for LRMs.

\begin{figure}[!t]
\centering
\includegraphics[width=\linewidth]{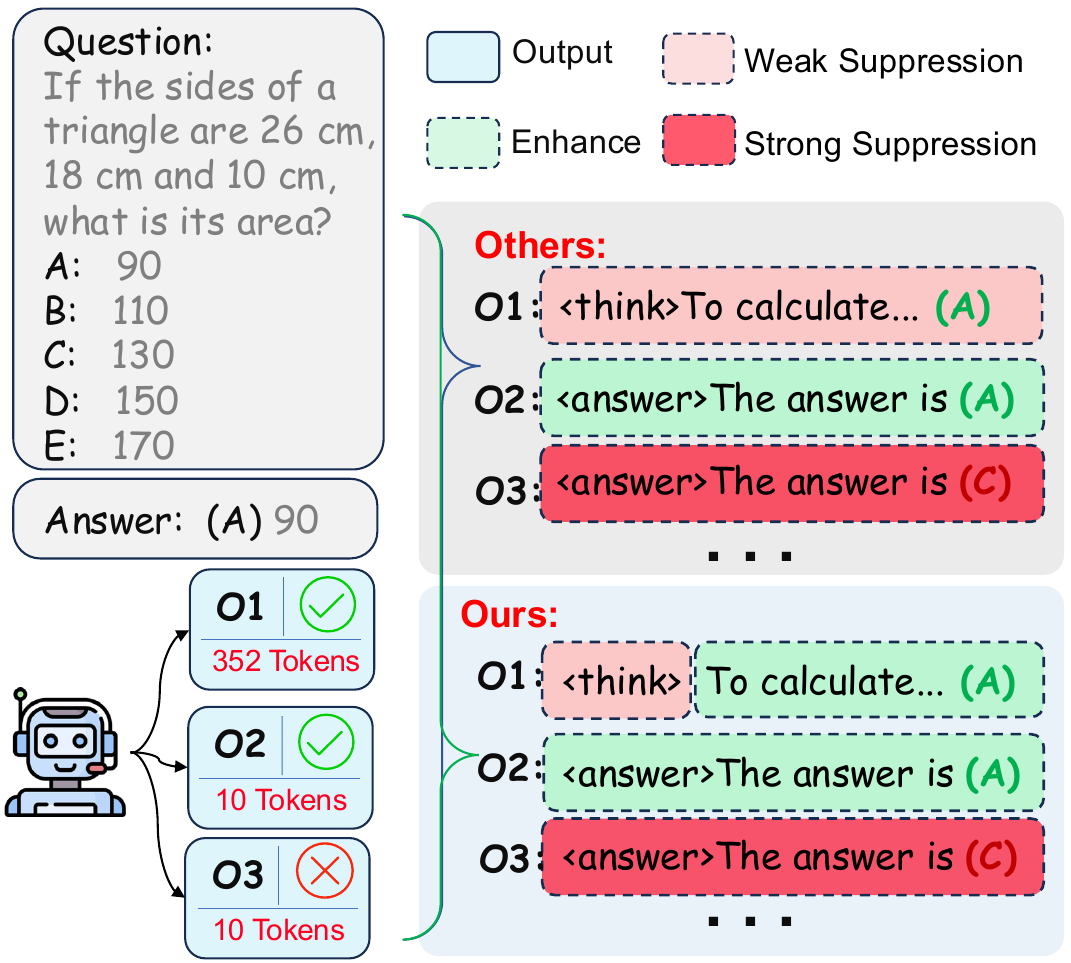}
\caption{In commonly used methods, efficiency penalties are applied at the sequence level, which penalizes correct but long reasoning and degrades reasoning performance.
In contrast, ADaPT applies efficiency penalties only to the \texttt{<think>} token that controls reasoning-mode selection, so correct long reasoning is not penalized simply for being long.
}
\label{fig:intro}
\vspace{-0.5em}
\end{figure}

To address this efficiency challenge, existing approaches attempt to reduce reasoning cost through two main directions: length compression methods and hybrid reasoning methods~\citep{Li2025DynamicMindAT, Wu2025ARMAR,He2025ThinkDialAO,Team2025KimiKS}.

However, these approaches often \textbf{improve efficiency at the expense of reasoning performance}, due to unresolved \textbf{conflicts between efficiency and correctness signals} during training. Length compression methods reward shorter outputs to improve efficiency, which creates a direct conflict with tasks that require sufficient reasoning depth. Once reasoning is shortened below a task-dependent threshold, performance inevitably degrades~\citep{Han2025YourMH,Lee2025HowWD}. 
Hybrid reasoning methods aim to adapt reasoning length to task difficulty~\citep{He2025ThinkDialAO,yang2025r}, but still rely on sequence-level efficiency rewards.
When multiple outputs achieve comparable correctness, shorter answers receive both correctness and efficiency rewards, whereas longer but correct reasoning receives only the correctness reward and is therefore placed at a relative disadvantage (Figure~\ref{fig:intro}). As a result, sequence-level efficiency training systematically suppresses correct long reasoning because it is less efficient, which ultimately leads to degraded deep reasoning capability.

To mitigate this signal conflict, we argue that \textbf{efficiency and correctness should be explicitly decoupled during training.} Based on this insight, we propose Adaptive Dual-Process Thinking (ADaPT), a token-level framework inspired by dual-process theory\citep{Evans2003-EVAITM-3} that explicitly models fast and slow reasoning and decouples efficiency signals from answer correctness during training. As shown in Figure~\ref{fig:intro}, ADaPT uses a mode-selection token to control reasoning modes, applying efficiency rewards only to this token while preserving correctness optimization.

By decoupling mode selection from answer correctness, ADaPT avoids penalizing correct long CoT reasoning simply for being long, thereby preserving deep reasoning capability while still encouraging efficient behavior when appropriate. Beyond mitigating performance degradation, ADaPT further enables precise control over the efficiency--performance trade-off at inference time.
After training, the probability of triggering the \texttt{<think>} token can be smoothly adjusted, allowing a single trained model to move continuously along the efficiency--performance Pareto frontier.

ADaPT adopts a two-stage training pipeline.
A supervised fine-tuning (SFT) stage equips the model with basic fast and slow reasoning behaviors, 
followed by a reinforcement learning stage based on a token-level variant of GRPO to optimize reasoning-mode selection.

The main contributions of this paper are as follows:
\begin{itemize}[nosep]
\item We identify that performance degradation in efficient reasoning methods primarily stems from sequence-level coupling between efficiency and correctness signals.
\item We propose ADaPT, a token-level dual-process framework that mitigates efficiency-induced performance degradation by explicitly decoupling these signals.
\item ADaPT enables precise and continuous inference time control over the efficiency--performance trade-off, allowing a single model to trace out a Pareto frontier.
\item Extensive experiments demonstrate that ADaPT significantly reduces inference cost while maintaining strong reasoning performance across multiple benchmarks.
\end{itemize}

%% file: Preliminary_v2.tex
\section{Preliminary}

\textbf{Group Relative Policy Optimization (GRPO).}
GRPO~\citep{Shao2024DeepSeekMathPT} has demonstrated strong performance across various tasks and can achieve efficient scalability within the RLVR paradigm.
It estimates advantage values by utilizing the reward scores of $N$ solutions sampled within the same query, eliminating the need for an additional value model.

Formally, let $\pi_{\theta_{\text{old}}}$ and $\pi_{\theta}$ represent the policy models before and after the update, respectively.
For a given problem $x$, a set of solution sequences $\{y^{(i)}\}_{i=1}^{N}$ sampled from $\pi_{\theta_{\text{old}}}$, and a reward function $R(\cdot)$, GRPO computes the advantage by normalizing rewards within the group:

\begin{equation}
\small
\label{grpo}
\begin{aligned}
\mathcal{J}_{\text{GRPO}}(\theta) &=
\mathbb{E}_{x \sim \mathcal{D}, \{y^{(i)}\}_{i=1}^{N} \sim \pi_{\theta_{\text{old}}}(\cdot|x)}
\frac{1}{N} \sum_{i=1}^{N}
\frac{1}{|y^{(i)}|} \sum_{t=0}^{|y^{(i)}|-1} \\
&\hspace{-1em}
\min \left[
\rho_{t}^{(i)}(\theta) \hat{A}_{t}^{(i)}, \quad
\tilde{\rho}_{t}^{(i)}(\theta) \hat{A}_{t}^{(i)}
\right],
\end{aligned}
\end{equation}

where $\rho_{t}^{(i)}(\theta)$ is the importance ratio, $\tilde{\rho}_{t}^{(i)}(\theta)$ is the clipped ratio, and $\hat{A}_{t}^{(i)}$ is the normalized advantage, defined as follows:
{
\small
\begin{align}
\rho_{t}^{(i)}(\theta) &= \frac{\pi_{\theta}(y_{t}^{(i)} \mid x, y_{<t}^{(i)})}{\pi_{\theta_{\text{old}}}(y_{t}^{(i)} \mid x, y_{<t}^{(i)})}, \label{eq:rho_def} \\
\tilde{\rho}_{t}^{(i)}(\theta) &= \text{clip}\left(\rho_{t}^{(i)}(\theta), 1 \pm \epsilon_{\text{clip}}\right), \label{eq:rho_tilde_def} \\
\hat{A}_{t}^{(i)} &= \frac{R(x, y^{(i)}) - \mu_R}{\sigma_R}, \label{eq:advantage_def}
\end{align}
}%
where $\mu_R$ and $\sigma_R$ denote the mean and standard deviation of rewards $\{R(x, y^{(j)})\}_{j=1}^{N}$ within the group.

%% file: method_v7.tex


\section{Method}

We introduce \textit{Adaptive Dual-Process Thinking} (ADaPT), a framework that enables large language models to adaptively choose between fast and slow reasoning while preserving deep reasoning capability.
The core idea of ADaPT is to explicitly model reasoning-mode selection via a dedicated mode token and to decouple efficiency optimization from answer correctness during training.
ADaPT is trained in two stages: (1) SFT stage, which exposes the model to both reasoning modes; (2) RL stage, which optimizes reasoning-mode selection using token-level rewards.

\subsection{Efficiency-Performance Conflict Analysis}
Before introducing ADaPT, we analyze why existing efficiency-oriented reasoning methods often suffer from systematic performance degradation.
Despite differences in design, these methods share a common root cause:
\textbf{efficiency and correctness are coupled at the sequence level}, leading to unavoidable training conflicts.

\paragraph{Why Existing Efficiency-Oriented Methods Fail.} For length-compression methods, this conflict is explicit. Length-related rewards scalarize efficiency and correctness into a single sequence-level objective, directly incentivizing shorter reasoning. When tasks require sufficient reasoning depth, this creates a seesaw effect between efficiency and accuracy, suppressing necessary intermediate steps and degrading performance on complex problems~\citep{Han2025YourMH,Lee2025HowWD,Han2024TokenBudgetAwareLR}. Hybrid reasoning methods instead rely on routing mechanisms to select among multiple reasoning modes. When trained with correctness-only sequence-level rewards, routing decisions cannot be reliably learned, causing models to default to uniformly slow reasoning to maximize accuracy~\citep{Wu2025ARMAR}.
To encourage mode diversity, prior work introduces additional efficiency-related signals such as length penalties~\citep{He2025ThinkDialAO,Wu2025ARMAR}. However, these signals remain sequence-level, coupling routing decisions with answer generation and implicitly penalizing longer but correct reasoning, often leading to unstable or collapsed routing behaviors.

\paragraph{A Structural View from RL.} From a reinforcement learning perspective, the fundamental issue with sequence-level efficiency-performance rewards is a \textbf{mismatch between what the reward measures and where the relevant decision is made}. A key observation is that efficiency depends only on the reasoning mode, not on the specific reasoning content generated within that mode. Once a mode is selected by the first action, subsequent actions can affect correctness but no longer influence efficiency.

In a standard MDP formulation\citep{feng2012dynamic}, the return is defined as the cumulative sum of future rewards along the trajectory.
The first action $a_1$, which corresponds to selecting the reasoning mode at the beginning of generation, is the only point at which the agent must trade off efficiency and expected task success, and thus its return should reflect both objectives.
In contrast, once the agent transitions to the subsequent state $s_2$, where the reasoning mode has already been fixed, the remaining return depends solely on whether the final answer is correct.
Subsequent actions should therefore optimize correctness only, since they cannot affect the efficiency outcome determined at the entry point. \textbf{This structural asymmetry implies that the sequence-level return should be decomposed accordingly.}

By contrast, sequence-level rewards that couple efficiency and correctness distribute efficiency pressure uniformly across all tokens, even though later actions have no control over the mode decision. This misalignment assigns efficiency-related signals to subsequent actions that cannot influence the reasoning mode, inducing a biased optimization objective that ultimately degrades performance under each fixed mode. \textbf{Decoupling efficiency into an early mode-selection reward} aligns the return with the decision structure and avoids this signal conflict.

\subsection{Stage 1: ADaPT-SFT}

SFT serves as a cold start to familiarize the model with predefined reasoning behaviors and their corresponding output formats.
Specifically, we define two reasoning modes:
(1) \texttt{<think>} mode, which generates a full chain-of-thought followed by the final answer;
(2) \texttt{<answer>} mode, which produces a short rationale or directly outputs the final answer.

During SFT, training examples are annotated with either \texttt{<think>} or \texttt{<answer>} formats based on task difficulty.
This allows the model to learn the basic behaviors associated with slow, explicit reasoning and fast, concise answering, providing a foundation for subsequent adaptive mode selection.

\subsection{Stage 2: ADaPT-GRPO}
After SFT, the model can generate multiple reasoning formats but cannot reliably select them based on task difficulty.
ADaPT-GRPO addresses this by: (1) Decoupling mode selection from answer correctness via token-level rewards. (2) Stabilizing mode learning with a balanced dual-start rollout.
This enables adaptive reasoning-mode selection without degrading long CoT reasoning capability.

\subsubsection{Token-Level Mode Reward.}

Standard GRPO optimizes correctness with sequence-level rewards, but applying mode preferences at the same level penalizes long yet correct reasoning. We therefore introduce a dedicated mode reward that decouples reasoning-mode selection from task correctness.

\paragraph{Mode Reward Design.}
In ADaPT-GRPO, we introduce a token-level mode reward applied to the \texttt{<think>} token:
\begin{equation}
\small
r_{\text{think}} =
- \big[\alpha (\mathrm{Acc}_{\text{a}} - \gamma)
+ (1-\alpha)(\mathrm{Acc}_{\text{a}} - \mathrm{Acc}_{\text{t}})\big],
\end{equation}
where $\mathrm{Acc}_{\text{a}}$ and $\mathrm{Acc}_{\text{t}}$ denote the accuracies of the \texttt{<answer>} and \texttt{<think>} modes, respectively, estimated from rollout data.

This reward consists of two complementary components.
The first term enforces an absolute quality threshold on fast reasoning: fast mode is considered reliable only when $\mathrm{Acc}_{\text{a}} > \gamma$, otherwise invoking \texttt{<think>} is encouraged.
The second term captures the relative advantage of slow reasoning, increasing the incentive to trigger \texttt{<think>} when it provides clear performance gains over fast reasoning.
The hyperparameter $\alpha$ balances these absolute and relative criteria.

During training, the mode reward explicitly ties the probability of emitting \texttt{<think>} to the reliability of fast reasoning, as measured by $\mathrm{Acc}_{\text{a}}$.
As a result, the learned policy internalizes a calibrated preference over invoking slow reasoning.
At inference time, this enables smooth control of reasoning depth by adjusting the generation threshold of the \texttt{<think>} token, without modifying model parameters.

\paragraph{Combine Rewards and Return Definition.}
In ADaPT-GRPO, we combine a token-level mode reward with the conventional sequence-level task reward. Specifically, for each prompt with $N$ sampled outputs $\{y^{(i)}\}_{i=1}^{N}$, the token-level reward is defined as:

\begin{equation}
\small
r_t =
\begin{cases}
r_{\text{think}}, & t=0,\ y_0 = \langle\text{think}\rangle, \\
r_{\text{seq}}, & t=T-1, \\
0, & \text{otherwise}.
\end{cases}
\end{equation}
The return \(G_t\) at each token is computed by summing future rewards:
\begin{equation}
\small
G_t = \sum_{k=t}^{T-1} r_k,
\end{equation}
which yields
\begin{equation}
\small
G_0 = r_{\text{think}} + R(x,y), \quad
G_t = R(x,y), \quad t > 0.
\end{equation}

\paragraph{Group-wise Global Normalization.}  
Due to the finer granularity of token-level rewards, the original normalization method in Eq.~\ref{eq:rho_tilde_def} cannot be directly applied.
We directly apply group-wise global normalization over all token-level returns within each prompt's rollout set, as in REINFORCE++~\citep{Hu2025REINFORCESC}.

Specifically, for a prompt with rollout set
\(
\mathcal{D}_{\text{group}} =
\{ G_t^{(i)} \mid i=1,\dots,N;\ t=0,\dots,T-1 \},
\)
the normalized token-level advantage is computed as:
\begin{equation}
\small
\tilde{A}_t^{(i)} =
\frac{G_t^{(i)} - \mathbb{E}[G]}
{\sqrt{\mathrm{Var}[G]} + \epsilon_{\text{norm}}},
\quad G \in \mathcal{D}_{\text{group}},
\end{equation}
where \(\mathbb{E}[\cdot]\) and \(\mathrm{Var}[\cdot]\) denote the return mean and variance over all tokens and rollouts in \(\mathcal{D}_{\text{group}}\), and \(\epsilon\) is a small constant for numerical stability.

\subsubsection{Balanced Dual-Start Rollout.}
\label{app:balanced_rollout_zhengwen}
Since ADaPT concentrates reasoning-mode selection into the first token, reliable learning at this position is critical.
To both prevent mode collapse and ensure stable optimization of the mode-selection policy, we adopt a balanced dual-start rollout strategy. 
Each rollout batch of size $n$ is evenly split into two groups: one forced to start with \texttt{<think>} and the other with \texttt{<answer>}, guaranteeing sufficient samples for both reasoning modes. The complete algorithmic description is deferred to Appendix~\ref{app:implementation}. 

Since the first token is explicitly controlled and follows a uniform behavior policy, we set the old-policy probability at the first step to
\begin{equation}
\pi_{\theta_{\text{old}}}(a_t \mid s_t) = \frac{1}{2}, \quad t = 0,
\end{equation}
when computing importance weights, thereby avoiding distribution mismatch.

\begin{figure}[!t]
    \centering
    \includegraphics[width=\linewidth]{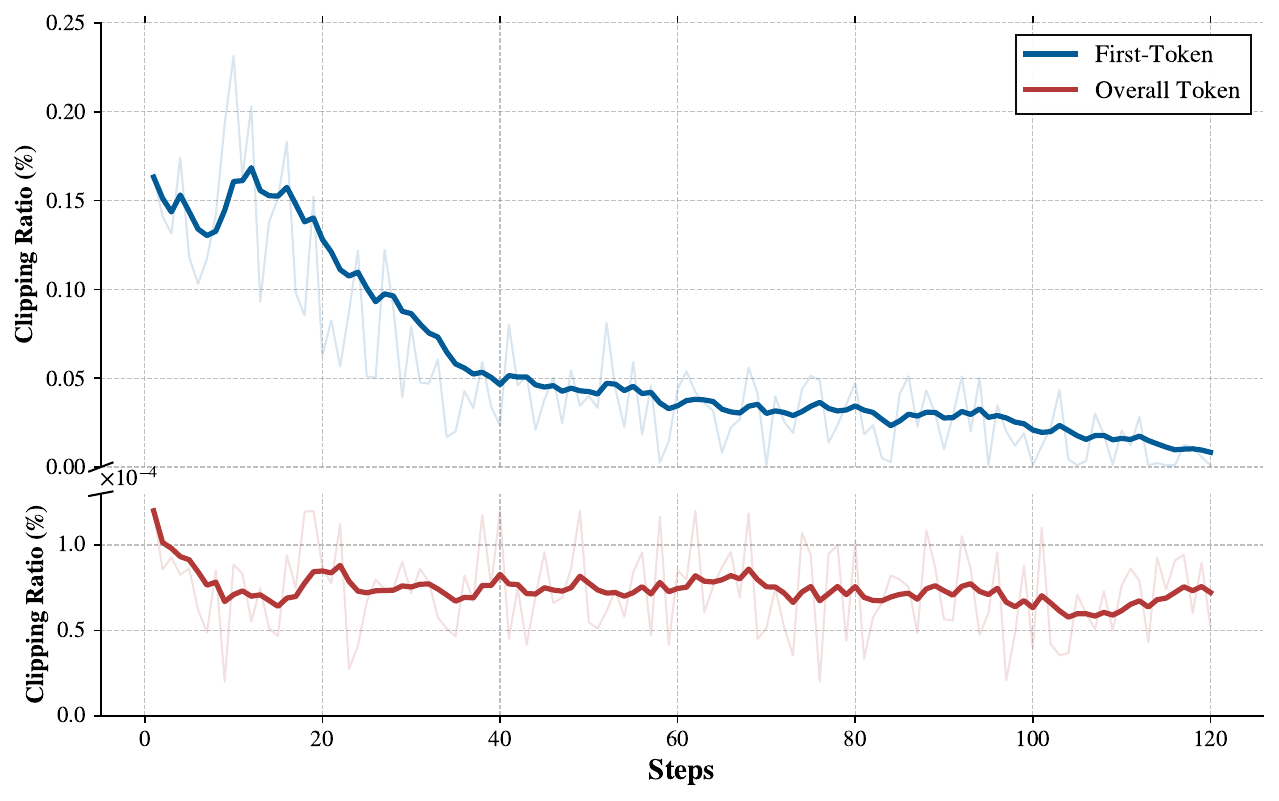}
    \caption{Clipping ratio during GRPO training. Clipping is rare for most tokens (bottom, $\times 10^{-4}$), but becomes much more frequent at the first token (top), which serves as the mode-selection decision point.}
    \label{fig:clip_ratio}
\end{figure}

However, applying standard PPO/GRPO clipping under this setting severely degrades learning at the first token, which serves as the mode-selection decision point. Due to the fixed behavior policy at $t=0$, the importance ratio is frequently clipped, leading to a much higher clipping rate at the mode-selection token than at other positions (Figure~\ref{fig:clip_ratio}). 

This excessive clipping effectively nullifies the policy gradient at the first token, resulting in vanishing or zero gradients for mode selection. Consequently, the model receives little to no learning signal to optimize when to trigger \texttt{<think>} versus \texttt{<answer>}, severely impairing its ability to learn reliable reasoning-mode selection.

To address this issue, we apply a CISPO~\cite{Chen2025MiniMaxM1ST} ratio constraint exclusively to the first token to ensure stable gradients, while all subsequent tokens are updated using the original GRPO/PPO objective.
The resulting token-level policy gradient objective is
\begin{equation}
\small
\mathcal{L}_{\text{PG}}^{\text{ADaPT}}(\theta)
= \sum_{i=1}^{N} \sum_{t=0}^{|y^{(i)}|-1}
L_{t}^{(i)\text{ADaPT}}(\theta),
\end{equation}

\begin{equation}
\small 
L_{t}^{(i)\text{ADaPT}}(\theta) =
\begin{cases}
\begin{aligned}
& - \text{detach}\big(\min(\rho_{0}^{(i)}, \epsilon_{\text{high}})\big) \\
& \quad \times \tilde{A}_{0}^{(i)} \log \pi_{\theta}(y_{0}^{(i)}),
\end{aligned}
& t = 0 \\
- \min \big(\rho_{t}^{(i)} \tilde{A}_{t}^{(i)},
\tilde{\rho}_{t}^{(i)} \tilde{A}_{t}^{(i)}\big),
& t \ge 1.
\end{cases}
\end{equation}

where $\rho_{t}^{(i)}$ and $\tilde{\rho}_{t}^{(i)}$ follow the definitions in Eq.~\ref{eq:rho_def} and Eq.~\ref{eq:rho_tilde_def}.
Finally, model parameters are optimized by maximizing
\begin{equation}
\small
\mathcal{J}_\text{ADaPT-GRPO}(\theta)
= \mathbb{E}\big[\mathcal{L}_\text{PG}^\text{ADaPT}(\theta)
- \beta D_\text{KL}[\pi_\theta \,\|\, \pi_\text{ref}]\big].
\end{equation}
\input{main_result}

%% file: main_result.tex
\begin{table*}[!t]
\centering

\def\arraystretch{0.99}
\setlength{\tabcolsep}{0.35em}

\resizebox{1.0\linewidth}{!}{
\begin{tabular}{l BB dd BB dd BB RR BB dd}
\toprule

\multirow{3}{*}{\textbf{Methods}}
& \multicolumn{6}{c}{\textbf{Easy}} 
& \multicolumn{8}{c}{\textbf{Hard}} 
& \multicolumn{2}{c}{\multirow{2}{*}{\textbf{AVG}}} \\

\cmidrule(lr){2-7} 
\cmidrule(lr){8-15}

& \multicolumn{2}{c}{\textbf{CSQA}}
& \multicolumn{2}{c}{\textbf{GSM8K}}
& \multicolumn{2}{c}{\textbf{ARC}}
& \multicolumn{2}{c}{\textbf{MATH500}}
& \multicolumn{2}{c}{\textbf{MMLU-Pro}}
& \multicolumn{2}{c}{\textbf{Olympiad}}
& \multicolumn{2}{c}{\textbf{AIME24}}
& \multicolumn{2}{c}{} \\ 

\cmidrule(lr){2-3}
\cmidrule(lr){4-5}
\cmidrule(lr){6-7}
\cmidrule(lr){8-9}
\cmidrule(lr){10-11}
\cmidrule(lr){12-13}
\cmidrule(lr){14-15}
\cmidrule(lr){16-17}

& ACC $\uparrow$ & Length $\downarrow$
& ACC $\uparrow$ & Length $\downarrow$
& ACC $\uparrow$ & Length $\downarrow$
& ACC $\uparrow$ & Length $\downarrow$
& ACC $\uparrow$ & Length $\downarrow$
& ACC $\uparrow$ & Length $\downarrow$
& ACC $\uparrow$ & Length $\downarrow$
& ACC $\uparrow$ & Length $\downarrow$ \\
\midrule

\rowcolor{bgcolor}
\multicolumn{17}{c}{\textbf{Qwen2.5-7B}} \\

Base
& 76.3 & 222
& 81.8 & 332
& 70.4 & 449
& 49.4 & 648
& 42.9 & 613
& 28.9 & 892
& 4.6 & 923
& 50.6 & 582 \\

SFT
& 79.4 & 376
& 68.3 & 541
& 73.2 & 949
& 44.0 & 632
& 40.1 & 893
& 24.8 & 1069
& 1.3 & 1484
& 47.3 & 849 \\

SFT+GRPO
& 84.3 & 391
& 91.1 & 801
& 87.8 & 405
& 77.4 & 1466
& 48.9 & 1007
& 39.5 & 2850
& 14.4 & 3857
& 63.3 & 1540 \\

TLMRE
& 82.8 & 204
& 88.3 & 584
& 82.6 & 287
& 75.4 & 1073
& 47.4 & 672
& 36.5 & 2782
& 14.2 & 3491
& 61.0 & 1299 \\

ARM
& 86.2 & 137
& 89.2 & 297
& 86.7 & 204
& 73.8 & 893
& 47.0 & 647
& 36.2 & 2362
& 13.1 & 3374
& 61.7 & 1131 \\

R-4B
& 85.7 & 145
& 88.6 & 310
& 84.7 & 274
& 74.4 & 912
& 46.9 & 771
& 34.9 & 2017
& 11.7 & 2883
& 60.9 & 1044 \\

\hdashline
$\mathrm{ADaPT}_{\text{answer}}$
& 86.1 & 24
& 84.7 & 152
& 85.3 & 42
& 67.2 & 427
& 44.4 & 317
& 31.7 & 795
& 6.3 & 1201
& 57.9 & 423 \\

$\mathrm{ADaPT}_{\text{think}}$
& 83.5 & 385
& 91.4 & 803
& 87.7 & 327
& 78.2 & 1541
& 49.8 & 1083
& 40.2 & 3071
& 15.1 & 3772
& 63.7 & 1569 \\

ADaPT
& 86.7 & 65
& 90.8 & 348
& 86.9 & 78
& 76.8 & 1003
& 48.5 & 680
& 35.8 & 1996
& 13.7 & 3044
& 62.7 & 1031 \\

\midrule
\rowcolor{bgcolor}
\multicolumn{17}{c}{\textbf{Qwen2.5-3B}} \\

Base
& 66.7 & 176
& 64.1 & 258
& 28.6 & 474
& 40.2 & 593
& 18.2 & 673
& 18.4 & 794
& 1.7 & 1103
& 33.9 & 582 \\

SFT
& 71.4 & 164
& 48.3 & 194
& 35.1 & 312
& 30.6 & 407
& 12.3 & 552
& 9.8 & 692
& 0.0 & 982
& 29.6 & 472 \\

SFT+GRPO
& 79.3 & 418
& 88.9 & 873
& 72.1 & 577
& 66.4 & 1483
& 29.4 & 1114
& 31.4 & 2873
& 4.3 & 3433
& 53.1 & 1539 \\

TLMRE
& 76.3 & 307
& 85.8 & 674
& 70.5 & 412
& 64.6 & 1178
& 28.4 & 1004
& 28.4 & 2085
& 3.5 & 3293
& 51.1 & 1279 \\

ARM
& 78.4 & 141
& 84.2 & 286
& 70.3 & 203
& 63.0 & 973
& 28.6 & 673
& 28.9 & 2341
& 3.9 & 3094
& 51.0 & 1101 \\

R-4B
& 76.9 & 226
& 83.6 & 427
& 68.3 & 184
& 62.8 & 1087
& 26.3 & 518
& 25.6 & 1864
& 2.5 & 2794
& 49.4 & 1014 \\

\hdashline
$\mathrm{ADaPT}_{\text{answer}}$
& 77.1 & 22
& 80.1 & 177
& 66.4 & 36
& 58.4 & 503
& 25.2 & 272
& 22.8 & 1033
& 2.0 & 1349
& 47.4 & 485 \\

$\mathrm{ADaPT}_{\text{think}}$
& 77.9 & 364
& 88.6 & 816
& 71.0 & 294
& 66.8 & 1492
& 27.5 & 1053
& 31.7 & 3289
& 4.6 & 3404
& 52.6 & 1530 \\

ADaPT
& 79.5 & 72
& 86.3 & 317
& 71.5 & 74
& 64.6 & 1074
& 28.2 & 583
& 29.5 & 2103
& 3.8 & 2872
& 51.9 & 1013 \\

\bottomrule
\end{tabular}
}

\caption{Performance comparison on Easy and Hard reasoning benchmarks.
Results report accuracy (ACC) and average generation length (Length) for different methods on two Qwen2.5 model scales. 
$ADaPT_{\text{think}}$ and $ADaPT_{\text{answer}}$ are constrained variants of ADaPT that force the model to always use slow reasoning or fast reasoning, respectively.}

\label{tab:main}
\end{table*}

%% file: experiment3.tex
\section{Experiment}

\subsection{Setup}
\paragraph{Datasets and Models.}
We conduct our experiments using the Qwen2.5 models~\citep{Yang2024Qwen25TR}, specifically the Qwen2.5-7B-Base and Qwen2.5-3B-Base variants.
The training data for the first stage is constructed entirely from publicly available datasets.
Specifically, we select a subset of the arm-team dataset\footnote{\url{https://huggingface.co/datasets/arm-team/Stage1_SFT_aqua_rat}} and utilize its long CoT and short CoT data to supervise the \texttt{<think>} and \texttt{<answer>} reasoning modes, respectively.
For the \texttt{<answer>} mode, we additionally include direct-answer samples when available.
In the second stage, we curate a total of 8.5k verifiable question–answer pairs from CSQA~\citep{Talmor2019CommonsenseQAAQ}, GSM8K~\citep{Cobbe2021TrainingVT}, and MATH~\citep{Hendrycks2021MeasuringMP}.
These tasks cover a wide range of difficulty levels, from relatively simple commonsense reasoning to complex mathematical problem solving.
Further experimental settings and dataset details are provided in Appendix~\ref{ap:exp}.

\paragraph{Evaluations.} We selected a diverse collection of datasets across commonsense reasoning, mathematics, and open-ended question answering. This selection includes both in-domain and out-of-domain test data, which are subsequently categorized into two difficulty groups. Easy Problems: CSQA, GSM8K, ARC~\cite{clark2018think}. Hard Problems: MATH500, MMLU-Pro~\citep{Wang2024MMLUProAM}, Olympiad~\citep{He2024OlympiadBenchAC}, and AIME24\footnote{\url{https://huggingface.co/datasets/Maxwell-Jia/AIME_2024}}. For AIME24, since the test set is relatively small, we report the avg@32 metric at a temperature of 0.6. For all other benchmarks, we report the pass@1 metric at a temperature of 0.

\paragraph{Baselines.}
We compare ADaPT with several representative baselines, including:
\textbf{Base}, \textbf{SFT}, \textbf{SFT + GRPO}, and several recent efficiency-oriented RL methods, including \textbf{TLMRE}~\citep{Arora2025TrainingLM},
\textbf{ARM}~\citep{Wu2025ARMAR},
and \textbf{R-4B}~\citep{yang2025r}.
Detailed descriptions of all baselines are provided in Appendix~\ref{app:baselines}.

\subsection{Main Results}

Table~\ref{tab:main} summarizes the results of ADaPT on two Qwen2.5 model scales compared with several baselines, across Easy and Hard reasoning benchmarks.
Easy tasks mainly involve basic reasoning and knowledge recall, while Hard tasks require multi-step reasoning and complex intermediate derivations.

\textbf{SFT+GRPO substantially improves reasoning accuracy but incurs high reasoning cost.}
Compared with Base and SFT, SFT+GRPO achieves notable accuracy gains, especially on Hard tasks, confirming the effectiveness of reinforcement learning for complex reasoning.
However, these gains are accompanied by a significant increase in generation length.
For example, on Qwen2.5-7B, SFT+GRPO produces over 2k tokens on Olympiad and more than 3k tokens on AIME24, with similar trends observed on Qwen2.5-3B.
This suggests that when training is dominated by correctness rewards, the model tends to adopt uniformly long reasoning trajectories to maximize accuracy.

\begin{figure*}[!t]
    \centering
    \includegraphics[width=\linewidth]{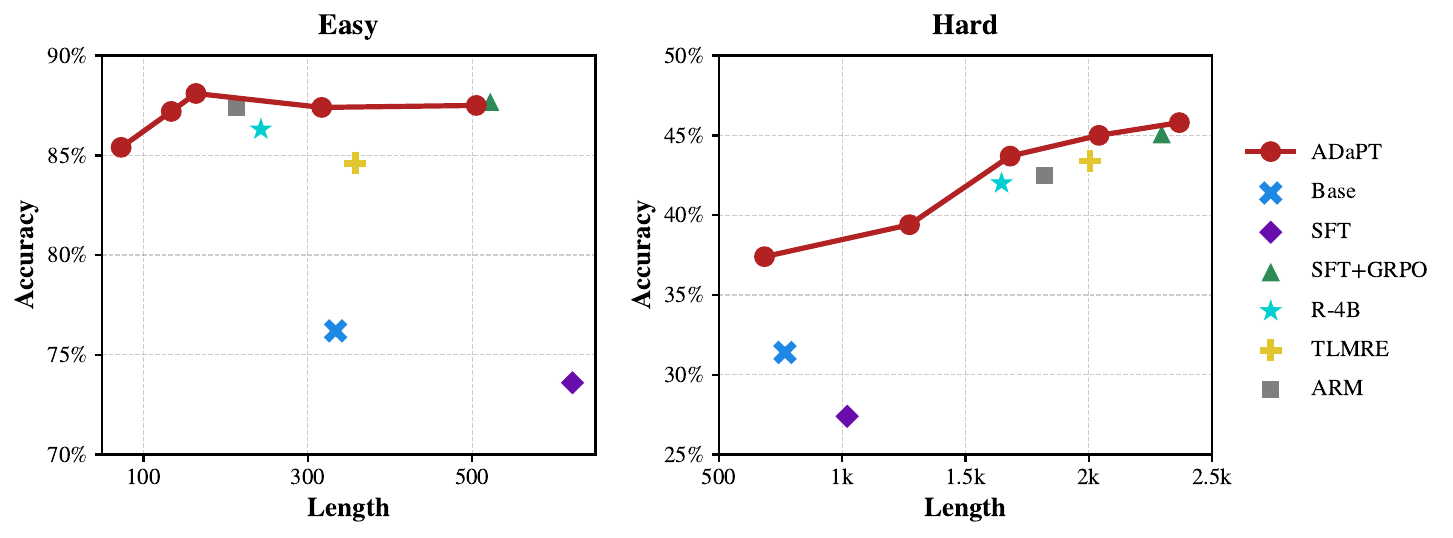}
    \caption{Accuracy and length trade-off of ADaPT on Easy (left) and Hard (right) tasks under different \texttt{<think>} token thresholds. By adjusting the threshold, ADaPT enables continuous Pareto control of reasoning efficiency, achieving Pareto optimal accuracy at different token budgets, while other methods remain inside this frontier.}

    \label{fig:optimized_pareto}
\end{figure*}

\textbf{ADaPT achieves a more favorable balance between performance and efficiency.}
Compared with SFT+GRPO, ADaPT significantly reduces generation length while incurring only marginal accuracy loss.
On Qwen2.5-7B, ADaPT reduces the average generation length from 1540 to 1031 tokens, with accuracy decreasing slightly from 63.1 to 62.7.
A similar pattern holds for Qwen2.5-3B.
In contrast, existing efficiency-oriented baselines such as TLMRE, ARM, and R-4B often reduce reasoning length at the cost of noticeable performance degradation, particularly on Hard tasks.
These results indicate that decoupling mode-selection rewards from correctness rewards enables the model to eliminate unnecessary reasoning without sacrificing accuracy.

\textbf{Different tasks exhibit distinct demands for fast and slow reasoning.}
On challenging benchmarks such as Olympiad and AIME24, forcing slow reasoning ($ADaPT_{\text{think}}$) substantially increases token usage but also yields clear accuracy gains.
Conversely, on simpler tasks like CSQA, slow reasoning leads to longer outputs with lower accuracy, while fast reasoning ($ADaPT_{\text{answer}}$) achieves competitive performance with minimal cost.
This highlights that indiscriminately increasing reasoning length can be harmful on simple problems.

\textbf{ADaPT preserves strong long-chain reasoning capability.}
Across both model scales, $ADaPT_{\text{think}}$ matches or slightly outperforms SFT+GRPO on most benchmarks.
For instance, on Qwen2.5-7B, $ADaPT_{\text{think}}$ achieves an average accuracy of 63.7, exceeding the 63.1 of SFT+GRPO.
Although slow reasoning remains costly, these results demonstrate that token-level mode selection rewards do not weaken deep reasoning ability, and the model retains the capacity to perform complex multi-step reasoning when required.

\subsection{Continuous Pareto Control of Reasoning Efficiency} \label{subsec:controllable_reasoning} A key property of ADaPT is that \textbf{reasoning depth can be continuously controlled by adjusting the generation probability threshold of the \texttt{<think>} token.} When the probability of generating \texttt{<think>} exceeds a given threshold, the model enters the slow-reasoning mode. We evaluate different threshold settings on Easy and Hard tasks and visualize the resulting accuracy–length trade-off in Figure~\ref{fig:optimized_pareto}.Across both task categories, increasing the threshold leads to more frequent slow reasoning, resulting in a monotonic increase in generation length and a corresponding improvement in accuracy. On Easy tasks, accuracy increases moderately before saturating, while on Hard tasks, higher reasoning depth yields consistent performance gains. These results show that ADaPT learns a continuous and controllable efficiency–capability trade-off rather than relying on a discrete switch between reasoning modes. Moreover, when projecting other baselines into the same accuracy–length space, most lie strictly inside the Pareto frontier defined by ADaPT. This indicates that ADaPT achieves higher accuracy at comparable or lower token budgets, confirming its systematic advantage in balancing reasoning efficiency and performance.

\begin{figure*}[!t]
    \centering
    \includegraphics[width=\linewidth]{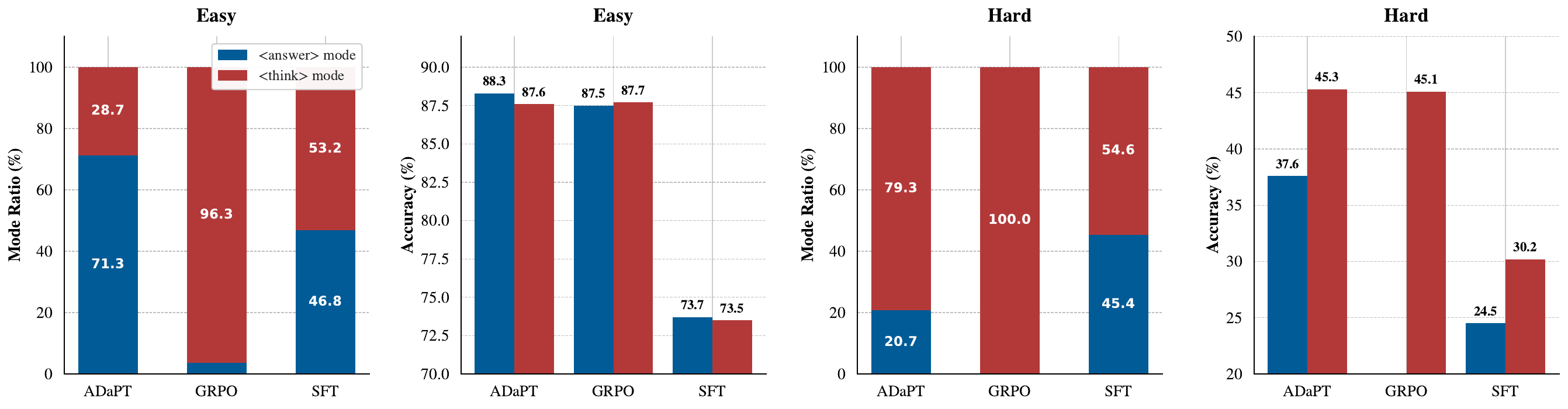}
    \caption{Distribution and accuracy of \texttt{<answer>} and \texttt{<think>} modes on Easy and Hard tasks. ADaPT adaptively selects fast reasoning for Easy tasks and increases slow reasoning only when required on Hard tasks, achieving a superior balance between efficiency and reasoning performance.}

    \label{fig:think_ratio}
\end{figure*}

\begin{figure}[!t]
    \centering
    \includegraphics[width=\linewidth]{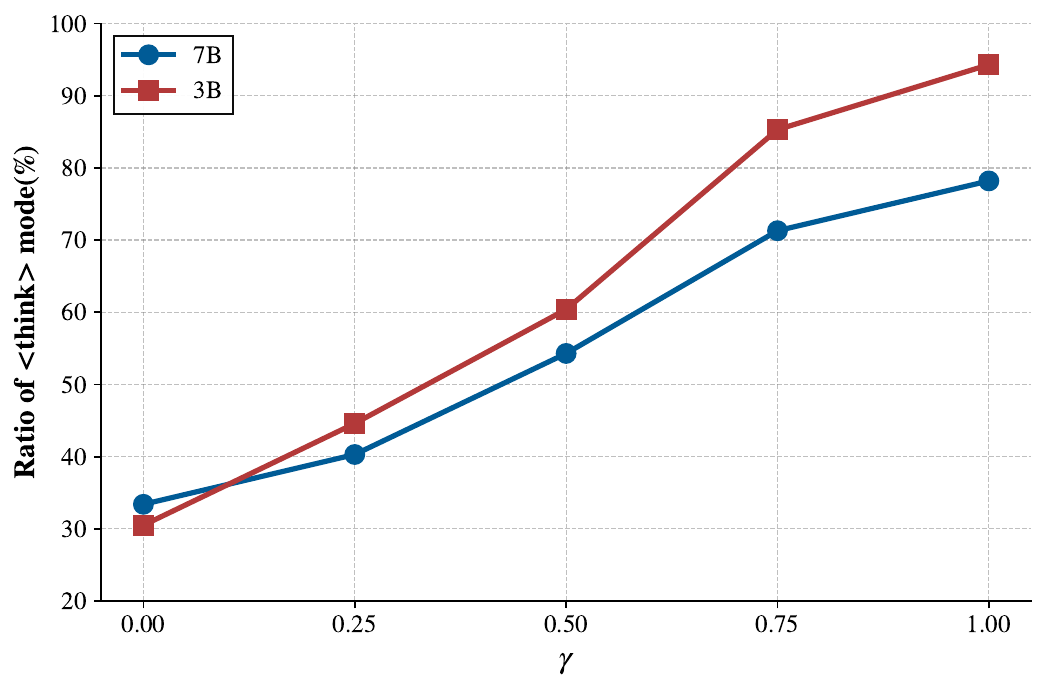}
    \caption{Effect of $\gamma$ on the proportion of slow reasoning (\texttt{<think>}) usage for 7B and 3B models. Larger $\gamma$ monotonically increases slow reasoning usage, with the 3B model showing a stronger shift due to lower fast reasoning reliability.}
    \label{fig:xiaorong}
\vspace{-0.5em}
\end{figure}


\subsection{Adaptive Use of Fast and Slow Reasoning}
Figure~\ref{fig:think_ratio} demonstrates that \textbf{ADaPT learns to select reasoning modes in a manner that is both accurate and sensitive to task difficulty.}
On Easy tasks, GRPO relies almost entirely on slow reasoning, reflecting training that is dominated by correctness rewards and therefore overuses long reasoning even when it is unnecessary.
SFT exposes the model to both reasoning formats, but lacks a mechanism to select between them, resulting in mixed usage with lower accuracy.
In contrast, ADaPT predominantly adopts fast reasoning on Easy tasks, achieving comparable accuracy with substantially lower inference cost.
On Hard tasks, GRPO again defaults to slow reasoning for all inputs, whereas ADaPT selectively increases the use of slow reasoning while preserving a meaningful proportion of fast reasoning.
Overall, ADaPT achieves strong accuracy under both reasoning modes and demonstrates effective and high performance reasoning mode selection across different task difficulties.

\subsection{Ablation Study on $\gamma$}
We ablate the parameter $\gamma$, which controls the tolerance of the mode reward to fast reasoning quality and regulates how easily the \texttt{<think>} mode is triggered.
As $\gamma$ increases, fast reasoning must satisfy a stricter reliability requirement, causing the model to invoke slow reasoning more frequently.
Figure~\ref{fig:xiaorong} shows a monotonic increase in \texttt{<think>} usage for both 7B and 3B models as $\gamma$ varies from 0 to 1, demonstrating that $\gamma$ enables smooth control over reasoning depth.
The 3B model exhibits a stronger response and surpasses the 7B model for $\gamma \ge 0.25$, consistent with its lower fast-reasoning reliability and greater reliance on slow reasoning as a fallback.

%% file: relatedwork.tex
\section{Related Work}
\subsection{Length Compression} 

Recently, many studies have focused on improving the reasoning efficiency of LLMs.
Some prompt-based approaches aim to simplify reasoning by modifying input prompts~\citep{Wu2025UnlockingEL,Muennighoff2025s1ST,Xu2025ChainOD,Chen2024DoNT}, for example by imposing explicit token budget constraints or instructing models to generate more concise reasoning chains~\citep{10852493,Han2024TokenBudgetAwareLR}.
Other methods emphasize early stopping strategies to reduce computational cost during reasoning~\citep{Han2025YourMH,Manvi2024AdaptiveIC,Li2024EscapeSC}.
In addition, several techniques reduce reasoning cost by explicitly controlling or pruning output length, typically through injecting multiple reasoning formats during pretraining or applying length penalties during reinforcement learning~\citep{Han2025YourMH,Team2025KimiKS,Shen2025DASTDS,Chang2025DemystifyingLC}.
Most of these approaches attempt to balance token budget and reasoning performance via explicit length constraints, often relying on accurate per-task token budget estimation or specialized training.

\subsection{Hybrid Reasoning} While many works improve efficiency by uniformly compressing reasoning length, hybrid reasoning offers a more flexible alternative by allowing models to adopt different reasoning strategies depending on the task~\citep{Fang2025ThinklessLL}. This paradigm requires models to switch between reasoning modes, which can be guided by external signals~\cite{Ong2024RouteLLMLT,Du2025CognitiveDR} or learned autonomously by the model itself~\citep{Fang2025ThinklessLL,Wu2025ARMAR,He2025ThinkDialAO,yang2025r,Zhang2025AdaptThinkRM}. Externally controlled approaches typically rely on routing mechanisms, such as complex control systems~\citep{Du2025CognitiveDR} or cascaded frameworks where a smaller model estimates task difficulty before invoking a larger one~\citep{Ong2024RouteLLMLT}. Other methods regulate reasoning behavior through explicit reasoning switches or prompts~\citep{Zhao2024AutomaticCE}. To enable autonomous reasoning-mode selection, many approaches rely on reinforcement learning. Some bias learning toward specific reasoning formats via sampling strategies~\citep{Wu2025ARMAR,yang2025r}, while others design reward functions to encourage more efficient reasoning~\citep{He2025ThinkDialAO,Zhang2025AdaptThinkRM}. However, when learning long and short reasoning simultaneously, these methods suffer from a fundamental limitation: sequence-level length rewards introduce unavoidable training conflicts between the two modes. 


%% file: conclusion.tex
\section{Conclusion}
In this work, we show that the commonly observed trade-off between reasoning efficiency and capability arises from sequence-level coupling of efficiency and correctness signals during training. To address this issue, we propose ADaPT, a token-level framework that decouples reasoning-mode selection from answer correctness via a dedicated mode-selection token. Experiments demonstrate that ADaPT substantially reduces inference cost while preserving strong long-chain reasoning ability, and enables continuous control over the efficiency–accuracy trade-off at inference time.

%% file: Limitation.tex
\section{Limitation}
This work has several limitations that point to directions for future research. First, our evaluation focuses on a set of standard reasoning benchmarks, mainly covering mathematical and commonsense reasoning. While these benchmarks span different difficulty levels, they may not fully reflect more diverse real-world settings, such as long-context or interactive reasoning. Second, ADaPT adopts a binary distinction between fast and slow reasoning modes for clarity and controllability. Although effective, this design does not capture more fine-grained variations in reasoning behavior, which could be explored in future extensions. Finally, experiments are conducted on models up to the 7B scale. While the approach is generally applicable, its behavior on larger models and under different training regimes remains to be investigated.

\section{Acknowledgments}
This work was supported by Ant Group.

%% file: appendix.tex
\section{Experiments Details}
\label{ap:exp}
\subsection{Hardware and Software Platform. }All experiments are conducted on workstations equipped with four NVIDIA A800 PCIe GPUs with 80GB memory each, running Ubuntu 20.04.6 LTS. Our implementation is based on the Verl~\citep{sheng2025hybridflow} framework. Rollouts are generated using temperature $= 0.6$. We set the maximum prompt length to 2,048 tokens and the maximum response length to 8192 tokens. Training is conducted for 120 steps with a batch size of 128 and a mini-batch size of 64. The actor is optimized using Adam, with learning rates of $1 \times 10^{-6}$. Our implementation uses grouped sampling with group size $G = 16$.

\subsection{Details of Training Data}
\paragraph{ADaPT-SFT Data Construction.}
For the SFT stage, we first use the base model to sample each question in the arm-team dataset ten times.
If the model answers a question correctly with a probability greater than 0.5, we classify the question as easy; otherwise, it is classified as hard.
For easy questions, we include both the standard chain-of-thought (CoT) data and the direct-answer data as training samples.
For hard questions, we include only the long CoT data as training samples.
This procedure establishes a difficulty-aware training curriculum and allows us to construct SFT data with explicit reasoning-depth supervision.
Finally, we select 5k easy questions and 5k hard questions, resulting in a total of 15k training instances, since each easy question corresponds to two reasoning formats.

\paragraph{Baseline Details}
\label{app:baselines}
We compare ADaPT with several representative baselines: (1) \textbf{Base}: The foundational model without any fine-tuning. (2) \textbf{SFT}: The model trained only through the first-stage SFT. This baseline is used to investigate whether SFT alone can instill the capability of accurate format selection in the model. (3) \textbf{SFT + GRPO}: The model is first trained with Stage 1 SFT and then further trained using the standard GRPO algorithm. This explores the final distribution of the two modes under a free-exploration setting. (4) \textbf{TLMRE}~\cite{Arora2025TrainingLM}: An RL algorithm that incorporates a length penalty to encourage the model to generate shorter responses. (5) \textbf{ARM}~\citep{Wu2025ARMAR}: A GRPO-based method designed to improve reasoning efficiency via four-mode routing. (6) \textbf{R-4B}~\citep{yang2025r}: An RL algorithm that teaches efficient thinking to models through bi-mode annealing.

\input{roolout}

\paragraph{ADaPT-GRPO Data Example.}
For the GRPO data, we select samples from CSQA, GSM8K, and MATH, with a total of 8.5k instances.
For each of these three datasets, we provide representative examples, as shown in Figure \ref{ap:GRPO_data}.

\section{Additional Implementation Details}
\label{app:implementation}

This appendix provides the detailed rollout procedure for the balanced dual-start strategy introduced in Section~\ref{app:balanced_rollout_zhengwen}, as summarized in Algorithm~\ref{alg:balanced_dual_start_rollout}.
The goal is to ensure sufficient and balanced sampling of both reasoning modes during training, while stabilizing learning at the mode-selection token.

\section{Selected Additional Results}
\label{app:additional}

\subsection{Cross-Backbone Generalization}
\label{app:cross_backbone}

To verify that ADaPT generalizes beyond the Qwen2.5 family, we apply the identical pipeline to LLaMA3-8B~\citep{Grattafiori2024TheL3}, which differs in tokenizer, pretraining corpus, and architecture.
All hyperparameters are kept the same as the Qwen2.5-7B setting.

\begin{table}[th]
\centering
\def\arraystretch{1.05}
\setlength{\tabcolsep}{0.45em}
\resizebox{1.0\linewidth}{!}{
\begin{tabular}{l l ccc}
\toprule
\textbf{Benchmark} & \textbf{Metric} & \textbf{Base} & \textbf{SFT+GRPO} & \textbf{ADaPT} \\
\midrule
\multirow{2}{*}{CSQA}     & ACC    & 25.3 & 57.4          & \textbf{58.8} \\
                           & Length & 52   & 231           & \textbf{64}   \\
\midrule
\multirow{2}{*}{GSM8K}    & ACC    & 21.3 & \textbf{50.8} & 50.1          \\
                           & Length & 273  & 514           & \textbf{321}  \\
\midrule
\multirow{2}{*}{MATH500}  & ACC    & 10.6 & \textbf{23.7} & 22.9          \\
                           & Length & 593  & 781           & \textbf{562}  \\
\midrule
\multirow{2}{*}{MMLU-Pro} & ACC    & 0.3  & \textbf{34.6} & 34.5          \\
                           & Length & 451  & 852           & \textbf{677}  \\
\midrule
\multirow{2}{*}{AIME24}   & ACC    & 0.1  & \textbf{0.4}  & \textbf{0.4}  \\
                           & Length & 942  & 1543          & \textbf{1231} \\
\bottomrule
\end{tabular}
}
\caption{Cross-backbone generalization results on LLaMA3-8B. ADaPT achieves comparable accuracy to SFT+GRPO while significantly reducing generation length across all benchmarks.}
\label{tab:cross_backbone}
\end{table}

As shown in Table~\ref{tab:cross_backbone}, ADaPT matches SFT+GRPO in accuracy while consistently shortening outputs (e.g., CSQA length drops from 231 to 64 tokens with a slight accuracy gain).
This confirms that token-level reward decoupling is architecture-agnostic and does not rely on Qwen-specific inductive biases.

\subsection{Scalability to Larger Models}
\label{app:scalability}

To examine whether ADaPT scales with model capacity, we apply it to Qwen2.5-14B~\citep{Yang2024Qwen25TR} and evaluate across all seven benchmarks.


\begin{table}[h]
\centering
\def\arraystretch{1.05}
\setlength{\tabcolsep}{0.45em}
\resizebox{1.0\linewidth}{!}{
\begin{tabular}{l l ccc}
\toprule
\textbf{Benchmark} & \textbf{Metric} & \textbf{Base} & \textbf{SFT+GRPO} & \textbf{ADaPT} \\
\midrule
\multirow{2}{*}{CSQA}     & ACC    & 78.3 & 85.1          & \textbf{85.8} \\
                           & Length & 67   & 512           & \textbf{97}   \\
\midrule
\multirow{2}{*}{GSM8K}    & ACC    & 82.1 & \textbf{93.2} & 93.1          \\
                           & Length & 173  & 577           & \textbf{164}  \\
\midrule
\multirow{2}{*}{ARC}      & ACC    & 74.1 & 90.1          & \textbf{90.4} \\
                           & Length & 495  & 675           & \textbf{89}   \\
\midrule
\multirow{2}{*}{MATH500}  & ACC    & 50.3 & \textbf{81.4} & 80.8          \\
                           & Length & 403  & 1693          & \textbf{995}  \\
\midrule
\multirow{2}{*}{MMLU-Pro} & ACC    & 44.2 & \textbf{51.6} & 51.1          \\
                           & Length & 673  & 2044          & \textbf{606}  \\
\midrule
\multirow{2}{*}{Olympiad} & ACC    & 30.1 & \textbf{42.8} & 40.9          \\
                           & Length & 832  & 3117          & \textbf{1964} \\
\midrule
\multirow{2}{*}{AIME24}   & ACC    & 5.1  & \textbf{18.7} & 18.3          \\
                           & Length & 1033 & 3782          & \textbf{3082} \\
\bottomrule
\end{tabular}
}
\caption{Scalability results on Qwen2.5-14B. ADaPT maintains comparable accuracy to SFT+GRPO while achieving substantial reductions in generation length.}
\label{tab:scalability}
\end{table}

As shown in Table~\ref{tab:scalability}, ADaPT's advantage persists and even grows at 14B scale: on easy benchmarks the length reductions reach 81\%--87\% (e.g., CSQA: 512$\rightarrow$97), while on harder ones accuracy stays within 1 point of SFT+GRPO with 41\%--70\% fewer tokens.
The consistent pattern across 3B, 7B, and 14B confirms that the method scales stably with model capacity.

\section{System Prompt and Case Studies}
We present our system prompt and example cases for the two reasoning modes below.

\clearpage
\input{GRPO_data}

\input{system_prompt}
\input{case_answer}

\input{case_think}

%% file: roolout.tex
\begin{algorithm}[t]
\caption{Balanced Dual-Start Rollout}
\label{alg:balanced_rollout_only}
\begin{algorithmic}[1]
\Require prompt $x$, policy $\pi_\theta$, batch size $N$, mode tokens $\tau_{\text{t}}=\texttt{<think>}$, $\tau_{\text{a}}=\texttt{<answer>}$
\Ensure $\mathcal{D}=\{(x,y^{(i)})\}_{i=1}^{N}$ with $N/2$ samples per mode
\State $\mathcal{D}\leftarrow\emptyset$
\For{$i=1$ \textbf{to} $N$}
  \State $y^{(i)}_{0}\leftarrow \mathbb{I}[i\le N/2]\tau_{\text{t}}+\mathbb{I}[i>N/2]\tau_{\text{a}}$
  \State $y^{(i)}_{1:T_i}\sim \prod_{t=1}^{T_i}\pi_\theta(\cdot\mid x,y^{(i)}_{0:t-1})$ \Comment{until \texttt{<eos>} / max length}
  \State $\mathcal{D}\leftarrow \mathcal{D}\cup\{(x,(y^{(i)}_{0},y^{(i)}_{1:T_i}))\}$
\EndFor
\State \Return $\mathcal{D}$
\end{algorithmic}
\label{alg:balanced_dual_start_rollout}
\end{algorithm}

%% file: GRPO_data.tex
\begin{figure*}[t]
\begin{tcolorbox}[title=GRPO Training data]
\begin{spacing}{1.3}
\textbf{\color{gray}{/*CSQA*/}} \\
\textbf{Question}: Where are you likely to find a professional prostitute?
\begin{enumerate}[label=\Alph*., itemsep=0pt, parsep=0pt, topsep=5pt] 
    \item new york
    \item whorehouse
    \item street corner
    \item corner of two streets
    \item brother
\end{enumerate}

\textbf{Answer}: B

\vspace{\medskipamount}
\vspace{\medskipamount}

\textbf{\color{gray}{/*GSM8K*/}} \\
\textbf{Question}: John buys 10 packs of magic cards.  Each pack has 20 cards and 1/4 of those cards are uncommon.  How many uncommon cards did he get?

\textbf{Answer}: 4
\vspace{\medskipamount}
\vspace{\medskipamount}

\textbf{\color{gray}{/*MATH*/}} \\
\textbf{Question}: To calculate $31^2$, Emily mentally figures the value $30^2$ and adds 61. Emily subtracts a number from $30^2$ to calculate $29^2$. What number does she subtract?

\textbf{Answer}: 59

\end{spacing}
\end{tcolorbox}

\caption{Examples of GRPO Training Data}
\label{ap:GRPO_data}
\end{figure*}

%% file: system_prompt.tex
\begin{figure*}[t]
\begin{tcolorbox}[title=System Prompt]
\begin{spacing}{1.3}
\textbf{\color{gray}{/*System Prompt*/}} \\
Please reason step by step. For difficult questions, output \textbf{\textcolor{purple}{<think>}} and engage long thinking mode. For simple questions, output \textbf{\textcolor{purple}{<answer>}} and engage short thinking mode. Provide your final answer within \verb|\boxed{}|.
\vspace{\medskipamount}
\vspace{\medskipamount}

\textbf{\color{gray}{/*User Prompt*/}} \\
Let $a > 0$, and let $P(x)$ be a polynomial with integer coefficients such that
\[
P(1) = P(3) = P(5) = P(7) = a
\]
and
\[
P(2) = P(4) = P(6) = P(8) = -a.
\]
What is the smallest possible value of $a$?

\end{spacing}
\end{tcolorbox}
\end{figure*}

%% file: case_answer.tex
\begin{figure*}[t]
\begin{tcolorbox}[title=Case Study 1: <answer> mode]
\begin{spacing}{1.3}

\textbf{\color{gray}{/*User Prompt*/}} \\
A basket contains 25 oranges among which 1 is bad, 20\% are unripe, 2 are sour and the rest are good. How many oranges are good?

\textbf{\color{gray}{/*Response*/}} \\
\textbf{\textcolor{purple}{<answer>}} \\
There are 25 oranges in total. One is bad, 20\% of them are unripe, which is 5 oranges, and 2 are sour. Subtracting these from the total gives 25 minus 1 minus 5 minus 2, which equals 17. Therefore, there are 17 good oranges.
The answer is \boxed{17}.

\vspace{\medskipamount}
\vspace{\medskipamount}
\vspace{\medskipamount}

\textbf{\color{gray}{/*User Prompt*/}} \\
What happens to water in a leaky faucet?
\begin{enumerate}[label=\Alph*., itemsep=0pt, parsep=0pt, topsep=5pt] 
    \item water spout
    \item wet clothes
    \item dribble
    \item power turbine
    \item thin soup
\end{enumerate}
\textbf{\color{gray}{/*Response*/}} \\
\textbf{\textcolor{purple}{<answer>}} \\
A leaky faucet allows water to drip continuously, which is a form of dribbling. The water does not form a spout, wet clothes, power a turbine, or become thin soup.\\
The answer is \boxed{C}.

\vspace{\medskipamount}
\vspace{\medskipamount}
\vspace{\medskipamount}

\textbf{\color{gray}{/*User Prompt*/}} \\
Snow, rain, hail, and fog are all forms of
\begin{enumerate}[label=\Alph*., itemsep=0pt, parsep=0pt, topsep=5pt] 
    \item gas
    \item water
    \item wind
    \item clouds
\end{enumerate}

\textbf{\color{gray}{/*Response*/}} \\
\textbf{\textcolor{purple}{<answer>}} \\
The answer is \boxed{B}.

\end{spacing}
\end{tcolorbox}
\end{figure*}

%% file: case_think.tex
\begin{figure*}[t]
\begin{tcolorbox}[title=Case Study 2: <think> mode]
\begin{spacing}{1.3}
\textbf{\color{gray}{/*User Prompt*/}} \\
Define
\[
p = \sum_{k = 1}^\infty \frac{1}{k^2} \quad \text{and} \quad q = \sum_{k = 1}^\infty \frac{1}{k^3}.
\]
Find a way to write
\[
\sum_{j = 1}^\infty \sum_{k = 1}^\infty \frac{1}{(j + k)^3}
\]
in terms of $p$ and $q$.
\vspace{\medskipamount}

\textbf{\color{gray}{/*Response*/}} \\
\textbf{\textcolor{purple}{<think>}} \\
Okay, so I have this problem here about these infinite series involving $p$ and $q$. Let me try to understand what I need to do. So, the problem says that $p$ is the sum of the reciprocals of the squares of the natural numbers, and $q$ is the sum of the reciprocals of the cubes. Then I need to find a way to express the double sum $\sum_{j=1}^\infty \sum_{k=1}^\infty \frac{1}{(j+k)^3}$ in terms of $p$ and $q$.

First, I remember that $p$ is the Riemann zeta function evaluated at 2, so $p = \zeta(2) = \frac{\pi^2}{6}$. And $q$ is $\zeta(3)$, which is a bit more complicated, but I don't think I need the exact value of $q$ here. The main thing is to find a way to relate the double sum to these two sums.

So, the double sum is $\sum_{j=1}^\infty \sum_{k=1}^\infty \frac{1}{(j+k)^3}$. I think I should start by changing the order of summation. When you have a double sum, sometimes changing the order can help. So, instead of summing over $j$ first and then $k$, I can sum over $j+k$ first. Let me see how that works.

Let me denote $n = j + k$. Then for each $n$, $j$ can be from 1 to $n-1$ and $k = n - j$. So the double sum becomes $\sum_{n=2}^\infty \sum_{j=1}^{n-1} \frac{1}{n^3}$. Because when $n = 2$, $j$ can be 1, when $n=3$, $j$ can be 1 or 2, and so on. So the inner sum is $\sum_{j=1}^{n-1} \frac{1}{n^3} = \frac{n-1}{n^3}$. Therefore, the double sum is $\sum_{n=2}^\infty \frac{n-1}{n^3}$.

Now, I can simplify $\frac{n-1}{n^3}$ as $\frac{1}{n^2} - \frac{1}{n^3}$. So the sum becomes $\sum_{n=2}^\infty \left( \frac{1}{n^2} - \frac{1}{n^3} \right)$.

I can split this into two separate sums: $\sum_{n=2}^\infty \frac{1}{n^2} - \sum_{n=2}^\infty \frac{1}{n^3}$. Now, I know that $\sum_{n=1}^\infty \frac{1}{n^2} = p = \frac{\pi^2}{6}$, so $\sum_{n=2}^\infty \frac{1}{n^2} = p - 1$. Similarly, $\sum_{n=1}^\infty \frac{1}{n^3} = q$, so $\sum_{n=2}^\infty \frac{1}{n^3} = q - 1$. Therefore, the sum becomes $(p - 1) - (q - 1) = p - 1 - q + 1 = p - q$.

So the answer should be $p - q$. Let me double-check to make sure I didn't make any mistakes. Let me go through the steps again.

Original double sum: $\sum_{j=1}^\infty \sum_{k=1}^\infty \frac{1}{(j+k)^3}$. By changing the order to sum over $n = j+k$, the inner sum is from $j=1$ to $n-1$, so $\sum_{j=1}^{n-1} \frac{1}{n^3} = \frac{n-1}{n^3} = \frac{1}{n^2} - \frac{1}{n^3}$. Then summing from $n=2$ to infinity gives $\sum_{n=2}^\infty \frac{1}{n^2} - \sum_{n=2}^\infty \frac{1}{n^3} = (p - 1) - (q - 1) = p - q$. Yeah, that seems right. \\
So the answer is \boxed{p - q}.

\end{spacing}
\end{tcolorbox}

\end{figure*}

%% file: ref.bib
@article{ElKishky2024OpenAIOS,
  title={OpenAI o1 System Card},
  author={Ahmed El-Kishky},
  journal={ArXiv},
  year={2024},
  volume={abs/2412.16720},
  url={https://api.semanticscholar.org/CorpusID:272648256}
}

@article{guo2025deepseek,
  title={Deepseek-r1: Incentivizing reasoning capability in llms via reinforcement learning},
  author={Guo, Daya and Yang, Dejian and Zhang, Haowei and Song, Junxiao and Zhang, Ruoyu and Xu, Runxin and Zhu, Qihao and Ma, Shirong and Wang, Peiyi and Bi, Xiao and others},
  journal={arXiv preprint arXiv:2501.12948},
  year={2025}
}

@article{Chen2025TowardsRE,
  title={Towards Reasoning Era: A Survey of Long Chain-of-Thought for Reasoning Large Language Models},
  author={Qiguang Chen and Libo Qin and Jinhao Liu and Dengyun Peng and Jiannan Guan and Peng Wang and Mengkang Hu and Yuhang Zhou and Te Gao and Wangxiang Che},
  journal={ArXiv},
  year={2025},
  volume={abs/2503.09567},
  url={https://api.semanticscholar.org/CorpusID:276937570}
}

@inproceedings{Zhang2025ASO,
  title={A Survey on Test-Time Scaling in Large Language Models: What, How, Where, and How Well?},
  author={Qiyuan Zhang and Fuyuan Lyu and Zexu Sun and Lei Wang and Weixu Zhang and Zhihan Guo and Yufei Wang and Irwin King and Xue Liu and Chen Ma},
  year={2025},
  url={https://api.semanticscholar.org/CorpusID:277467322}
}

@article{Luo2025O1PrunerLF,
  title={O1-Pruner: Length-Harmonizing Fine-Tuning for O1-Like Reasoning Pruning},
  author={Haotian Luo and Li Shen and Haiying He and Yibo Wang and Shiwei Liu and Wei Li and Naiqiang Tan and Xiaochun Cao and Dacheng Tao},
  journal={ArXiv},
  year={2025},
  volume={abs/2501.12570},
  url={https://api.semanticscholar.org/CorpusID:275790112}
}

@article{Wu2025ARMAR,
  title={ARM: Adaptive Reasoning Model},
  author={Siye Wu and Jian Xie and Yikai Zhang and Aili Chen and Kai Zhang and Yu Su and Yanghua Xiao},
  journal={ArXiv},
  year={2025},
  volume={abs/2505.20258},
  url={https://api.semanticscholar.org/CorpusID:278912115}
}

@article{Liu2025ThoughtME,
  title={Thought Manipulation: External Thought Can Be Efficient for Large Reasoning Models},
  author={Yule Liu and Jingyi Zheng and Zhen Sun and Zifan Peng and Wenhan Dong and Zeyang Sha and Shiwen Cui and Weiqiang Wang and Xinlei He},
  journal={ArXiv},
  year={2025},
  volume={abs/2504.13626},
  url={https://api.semanticscholar.org/CorpusID:277940127}
}

@inproceedings{Hashemi2025DNRBB,
  title={DNR Bench: Benchmarking Over-Reasoning in Reasoning LLMs},
  author={Masoud Hashemi and Oluwanifemi Bamgbose and Sathwik Tejaswi Madhusudhan and Jishnu Sethumadhavan Nair and Aman Tiwari and Vikas Yadav},
  year={2025},
  url={https://api.semanticscholar.org/CorpusID:277150831}
}

@article{Han2025YourMH,
  title={Your Models Have Thought Enough: Training Large Reasoning Models to Stop Overthinking},
  author={Jinyi Han and Ying Huang and Ying Liao and Zishang Jiang and Xikun Lu and Haiquan Zhao and Xinyi Wang and Guanghao Zhou and Sihang Jiang and Jiaqing Liang and Weikang Zhou and Zeye Sun and Fei Yu and Yanghua Xiao},
  journal={ArXiv},
  year={2025},
  volume={abs/2509.23392},
  url={https://api.semanticscholar.org/CorpusID:281675053}
}

@article{Fang2025ThinklessLL,
  title={Thinkless: LLM Learns When to Think},
  author={Gongfan Fang and Xinyin Ma and Xinchao Wang},
  journal={ArXiv},
  year={2025},
  volume={abs/2505.13379},
  url={https://api.semanticscholar.org/CorpusID:278769001}
}

@article{Li2025DynamicMindAT,
  title={DynamicMind: A Tri-Mode Thinking System for Large Language Models},
  author={Wei Li and Yanbin Wei and Qiushi Huang and Jiangyue Yan and Yang Chen and James T. Kwok and Yu Zhang},
  journal={ArXiv},
  year={2025},
  volume={abs/2506.05936},
  url={https://api.semanticscholar.org/CorpusID:279243931}
}

@article{Zhang2025AdaptThinkRM,
  title={AdaptThink: Reasoning Models Can Learn When to Think},
  author={Jiajie Zhang and Nianyi Lin and Lei Hou and Ling Feng and Juanzi Li},
  journal={ArXiv},
  year={2025},
  volume={abs/2505.13417},
  url={https://api.semanticscholar.org/CorpusID:278769267}
}

@article{Team2025KimiKS,
  title={Kimi k1.5: Scaling Reinforcement Learning with LLMs},
  author={Kimi Team and Angang Du and Bofei Gao and Bowei Xing and Changjiu Jiang and Cheng Chen and Cheng Li and Chenjun Xiao and Chenzhuang Du and Chonghua Liao and Chuning Tang and Congcong Wang and Dehao Zhang and Enming Yuan and Enzhe Lu and Feng Tang and Flood Sung and Guangda Wei and Guokun Lai and Haiqing Guo and Han Zhu and Haochen Ding and Hao-Xing Hu and Haoming Yang and Hao Zhang and Haotian Yao and Hao-Dong Zhao and Haoyu Lu and Haoze Li and Haozhen Yu and Hongcheng Gao and Huabin Zheng and Huan Yuan and Jia Chen and Jia-Xing Guo and Jianling Su and Jianzhou Wang and Jie Zhao and Jin Zhang and Jingyuan Liu and Junjie Yan and Junyan Wu and Li-Na Shi and Li-Tao Ye and Long Yu and Meng-xiao Dong and Neo Y. Zhang and Ningchen Ma and Qi Pan and Qucheng Gong and Shaowei Liu and Shen Ma and Shu-Yan Wei and Sihan Cao and Si-Da Huang and Tao Jiang and Wei-Wei Gao and Weiming Xiong and Weiran He and Weixiao Huang and Wenhao Wu and Wen He and Xian-sen Wei and Xian-Xian Jia and Xingzhe Wu and Xinran Xu and Xinxing Zu and Xinyu Zhou and Xue-biao Pan and Y. Charles and Yang Li and Yan-Ling Hu and Yangyang Liu and Yanru Chen and Ye-Jia Wang and Yibo Liu and Yidao Qin and Yifeng Liu and Yingbo Yang and Yiping Bao and Yulun Du and Yuxin Wu and Yuzhi Wang and Zaida Zhou and Zhaoji Wang and Zhaowei Li and Zhengxin Zhu and Zheng Zhang and Zhexu Wang and Zhilin Yang and Zhiqi Huang and Zihao Huang and Ziya Xu and Zonghan Yang},
  journal={ArXiv},
  year={2025},
  volume={abs/2501.12599},
  url={https://api.semanticscholar.org/CorpusID:275789974}
}

@article{He2025ThinkDialAO,
  title={ThinkDial: An Open Recipe for Controlling Reasoning Effort in Large Language Models},
  author={Qianyu He and Siyu Yuan and Xuefeng Li and Mingxuan Wang and Jiangjie Chen},
  journal={ArXiv},
  year={2025},
  volume={abs/2508.18773},
  url={https://api.semanticscholar.org/CorpusID:280869962}
}

@article{yang2025r,
  title={R-4b: Incentivizing general-purpose auto-thinking capability in mllms via bi-mode annealing and reinforce learning},
  author={Yang, Qi and Ni, Bolin and Xiang, Shiming and Hu, Han and Peng, Houwen and Jiang, Jie},
  journal={arXiv preprint arXiv:2508.21113},
  year={2025}
}

@article{Wu2025UnlockingEL,
  title={Unlocking Efficient Long-to-Short LLM Reasoning with Model Merging},
  author={Han Wu and Yuxuan Yao and Shuqi Liu and Zehua Liu and Xiaojin Fu and Xiongwei Han and Xing Li and Hui-Ling Zhen and Tao Zhong and Mingxuan Yuan},
  journal={ArXiv},
  year={2025},
  volume={abs/2503.20641},
  url={https://api.semanticscholar.org/CorpusID:277322544}
}

@article{Muennighoff2025s1ST,
  title={s1: Simple test-time scaling},
  author={Niklas Muennighoff and Zitong Yang and Weijia Shi and Xiang Lisa Li and Fei-Fei Li and Hanna Hajishirzi and Luke S. Zettlemoyer and Percy Liang and Emmanuel J. Cand{\`e}s and Tatsunori Hashimoto},
  journal={ArXiv},
  year={2025},
  volume={abs/2501.19393},
  url={https://api.semanticscholar.org/CorpusID:276079693}
}

@article{Xu2025ChainOD,
  title={Chain of Draft: Thinking Faster by Writing Less},
  author={Silei Xu and Wenhao Xie and Lingxiao Zhao and Pengcheng He},
  journal={ArXiv},
  year={2025},
  volume={abs/2502.18600},
  url={https://api.semanticscholar.org/CorpusID:276618268}
}

@INPROCEEDINGS{10852493,
  author={Renze, Matthew and Guven, Erhan},
  booktitle={2024 2nd International Conference on Foundation and Large Language Models (FLLM)}, 
  title={The Benefits of a Concise Chain of Thought on Problem-Solving in Large Language Models}, 
  year={2024},
  volume={},
  number={},
  pages={476-483},
  doi={10.1109/FLLM63129.2024.10852493}}

@article{Han2024TokenBudgetAwareLR,
  title={Token-Budget-Aware LLM Reasoning},
  author={Tingxu Han and Zhenting Wang and Chunrong Fang and Shiyun Zhao and Shiqing Ma and Zhenyu Chen},
  journal={ArXiv},
  year={2024},
  volume={abs/2412.18547},
  url={https://api.semanticscholar.org/CorpusID:274992044}
}

@article{Shen2025DASTDS,
  title={DAST: Difficulty-Adaptive Slow-Thinking for Large Reasoning Models},
  author={Yi Shen and Jian Zhang and Jie-fu Huang and Shuming Shi and Wenjing Zhang and Jiangze Yan and Ning Wang and Kai Wang and Shiguo Lian},
  journal={ArXiv},
  year={2025},
  volume={abs/2503.04472},
  url={https://api.semanticscholar.org/CorpusID:276813665}
}

@article{Chang2025DemystifyingLC,
  title={Demystifying Long Chain-of-Thought Reasoning in LLMs},
  author={Edward Y. Chang and Yuxuan Tong and Morry Niu and Graham Neubig and Xiang Yue},
  journal={ArXiv},
  year={2025},
  volume={abs/2502.03373},
  url={https://api.semanticscholar.org/CorpusID:276116814}
}

@article{Ong2024RouteLLMLT,
  title={RouteLLM: Learning to Route LLMs with Preference Data},
  author={Isaac Ong and Amjad Almahairi and Vincent Wu and Wei-Lin Chiang and Tianhao Wu and Joseph Gonzalez and Mohammed Waleed Kadous and Ion Stoica},
  journal={ArXiv},
  year={2024},
  volume={abs/2406.18665},
  url={https://api.semanticscholar.org/CorpusID:270764307}
}

@article{Du2025CognitiveDR,
  title={Cognitive Decision Routing in Large Language Models: When to Think Fast, When to Think Slow},
  author={Y. Du and C. Guo and W. Wang and G. Tang},
  journal={ArXiv},
  year={2025},
  volume={abs/2508.16636},
  url={https://api.semanticscholar.org/CorpusID:280710800}
}

@article{Shao2024DeepSeekMathPT,
  title={DeepSeekMath: Pushing the Limits of Mathematical Reasoning in Open Language Models},
  author={Zhihong Shao and Peiyi Wang and Qihao Zhu and Runxin Xu and Jun-Mei Song and Mingchuan Zhang and Y. K. Li and Yu Wu and Daya Guo},
  journal={ArXiv},
  year={2024},
  volume={abs/2402.03300},
  url={https://api.semanticscholar.org/CorpusID:267412607}
}

@inproceedings{Hu2025REINFORCESC,
  title={REINFORCE++: Stabilizing Critic-Free Policy Optimization with Global Advantage Normalization},
  author={Jian Hu and Jason Klein Liu and Haotian Xu and Wei Shen},
  year={2025},
  url={https://api.semanticscholar.org/CorpusID:282911716}
}

@article{Chen2025MiniMaxM1ST,
  title={MiniMax-M1: Scaling Test-Time Compute Efficiently with Lightning Attention},
  author={Aili Chen and Aonian Li and Bangwei Gong and Binyang Jiang and Bo Fei and Bo Yang and Boji Shan and Changqing Yu and Chao Wang and Cheng Zhu and Chengjun Xiao and Chengyu Du and Chi Zhang and Chu Qiao and Chunhao Zhang and Chunhui Du and Congchao Guo and Da Chen and Deming Ding and Dianjun Sun and Dong Li and Enwei Jiao and Haigang Zhou and Haimo Zhang and Han Ding and Haohai Sun and Haoyu Feng and Huaiguang Cai and Haichao Zhu and Jian Sun and Jiaqi Zhuang and Jia-Yao Cai and Jiayuan Song and Jin Zhu and Jingyang Li and Jinhao Tian and Jinli Liu and Junhao Xu and Junjie Yan and Junteng Liu and Junxia He and Kaiyi Feng and Ke Yang and Ke Xiao and Le Han and Leyang Wang and Lian-Chun Yu and Li Feng and Lin Li and Lin Zheng and Linge Du and Li-na Yang and Lunbin Zeng and Ming-Yuan Yu and Mingliang Tao and Mingyuan Chi and Mozhi Zhang and Mujie Lin and Nan Hu and Nongyu Di and Peng Gao and Pengfei Li and Peng Zhao and Qibing Ren and Qidi Xu and Qile Li and Qin Wang and Rong Tian and Ruitao Leng and Shaoxiang Chen and Shaoyu Chen and Shengmin Shi and Shitong Weng and Shuchang Guan and Shu Yu and Si-Jie Li and Song He Zhu and Tengfei Li and Tian-Yi Cai and Tianrun Liang and Weiyu Cheng and Weize Kong and Wenkai Li and Xian Feng Chen and Xiangjun Song and Xiao Luo and Xiaoyan Su and Xiaobo Li and Xiaodong Han and Xi-Yong Hou and Xuan-wen Lu and Xun Zou and Xuyang Shen and Yanting Gong and Yan Ma and Yang Wang and Yiqi Shi and Yi-Fan Zhong and Yonghong Duan},
  journal={ArXiv},
  year={2025},
  volume={abs/2506.13585},
  url={https://api.semanticscholar.org/CorpusID:279402731}
}

@article{Yang2024Qwen25TR,
  title={Qwen2.5 Technical Report},
  author={Qwen An Yang and Baosong Yang and Beichen Zhang and Binyuan Hui and Bo Zheng and Bowen Yu and Chengyuan Li and Dayiheng Liu and Fei Huang and Guanting Dong and Haoran Wei and Huan Lin and Jian Yang and Jianhong Tu and Jianwei Zhang and Jianxin Yang and Jiaxin Yang and Jingren Zhou and Junyang Lin and Kai Dang and Keming Lu and Keqin Bao and Kexin Yang and Le Yu and Mei Li and Mingfeng Xue and Pei Zhang and Qin Zhu and Rui Men and Runji Lin and Tianhao Li and Tingyu Xia and Xingzhang Ren and Xuancheng Ren and Yang Fan and Yang Su and Yi-Chao Zhang and Yunyang Wan and Yuqi Liu and Zeyu Cui and Zhenru Zhang and Zihan Qiu and Shanghaoran Quan and Zekun Wang},
  journal={ArXiv},
  year={2024},
  volume={abs/2412.15115},
  url={https://api.semanticscholar.org/CorpusID:274859421}
}

@article{Hendrycks2021MeasuringMP,
  title={Measuring Mathematical Problem Solving With the MATH Dataset},
  author={Dan Hendrycks and Collin Burns and Saurav Kadavath and Akul Arora and Steven Basart and Eric Tang and Dawn Xiaodong Song and Jacob Steinhardt},
  journal={ArXiv},
  year={2021},
  volume={abs/2103.03874},
  url={https://api.semanticscholar.org/CorpusID:232134851}
}

@article{Cobbe2021TrainingVT,
  title={Training Verifiers to Solve Math Word Problems},
  author={Karl Cobbe and Vineet Kosaraju and Mo Bavarian and Mark Chen and Heewoo Jun and Lukasz Kaiser and Matthias Plappert and Jerry Tworek and Jacob Hilton and Reiichiro Nakano and Christopher Hesse and John Schulman},
  journal={ArXiv},
  year={2021},
  volume={abs/2110.14168},
  url={https://api.semanticscholar.org/CorpusID:239998651}
}

@article{Talmor2019CommonsenseQAAQ,
  title={CommonsenseQA: A Question Answering Challenge Targeting Commonsense Knowledge},
  author={Alon Talmor and Jonathan Herzig and Nicholas Lourie and Jonathan Berant},
  journal={ArXiv},
  year={2019},
  volume={abs/1811.00937},
  url={https://api.semanticscholar.org/CorpusID:53296520}
}

@article{Wang2024MMLUProAM,
  title={MMLU-Pro: A More Robust and Challenging Multi-Task Language Understanding Benchmark},
  author={Yubo Wang and Xueguang Ma and Ge Zhang and Yuansheng Ni and Abhranil Chandra and Shiguang Guo and Weiming Ren and Aaran Arulraj and Xuan He and Ziyan Jiang and Tianle Li and Max W.F. Ku and Kai Wang and Alex Zhuang and Rongqi "Richard" Fan and Xiang Yue and Wenhu Chen},
  journal={ArXiv},
  year={2024},
  volume={abs/2406.01574},
  url={https://api.semanticscholar.org/CorpusID:270210486}
}

@article{clark2018think,
title={Think you have solved question answering? try arc, the ai2 reasoning challenge},
author={Clark, Peter and Cowhey, Isaac and Etzioni, Oren and Khot, Tushar and Sabharwal, Ashish and Schoenick, Carissa and Tafjord, Oyvind},
journal={arXiv preprint arXiv:1803.05457},
year={2018}
}

@inproceedings{He2024OlympiadBenchAC,
  title={OlympiadBench: A Challenging Benchmark for Promoting AGI with Olympiad-Level Bilingual Multimodal Scientific Problems},
  author={Chaoqun He and Renjie Luo and Yuzhuo Bai and Shengding Hu and Zhen Leng Thai and Junhao Shen and Jinyi Hu and Xu Han and Yujie Huang and Yuxiang Zhang and Jie Liu and Lei Qi and Zhiyuan Liu and Maosong Sun},
  booktitle={Annual Meeting of the Association for Computational Linguistics},
  year={2024},
  url={https://api.semanticscholar.org/CorpusID:267770504}
}

@article{Arora2025TrainingLM,
  title={Training Language Models to Reason Efficiently},
  author={Daman Arora and Andrea Zanette},
  journal={ArXiv},
  year={2025},
  volume={abs/2502.04463},
  url={https://api.semanticscholar.org/CorpusID:276235717}
}

@article{Chen2024DoNT,
  title={Do NOT Think That Much for 2+3=? On the Overthinking of o1-Like LLMs},
  author={Xingyu Chen and Jiahao Xu and Tian Liang and Zhiwei He and Jianhui Pang and Dian Yu and Linfeng Song and Qiuzhi Liu and Mengfei Zhou and Zhuosheng Zhang and Rui Wang and Zhaopeng Tu and Haitao Mi and Dong Yu},
  journal={ArXiv},
  year={2024},
  volume={abs/2412.21187},
  url={https://api.semanticscholar.org/CorpusID:275133600}
}

@article{Zhao2024AutomaticCE,
  title={Automatic Curriculum Expert Iteration for Reliable LLM Reasoning},
  author={Zirui Zhao and Hanze Dong and Amrita Saha and Caiming Xiong and Doyen Sahoo},
  journal={ArXiv},
  year={2024},
  volume={abs/2410.07627},
  url={https://api.semanticscholar.org/CorpusID:273233951}
}

@article{Manvi2024AdaptiveIC,
  title={Adaptive Inference-Time Compute: LLMs Can Predict if They Can Do Better, Even Mid-Generation},
  author={Rohin Manvi and Anikait Singh and Stefano Ermon},
  journal={ArXiv},
  year={2024},
  volume={abs/2410.02725},
  url={https://api.semanticscholar.org/CorpusID:273098836}
}

@article{Li2024EscapeSC,
  title={Escape Sky-high Cost: Early-stopping Self-Consistency for Multi-step Reasoning},
  author={Yiwei Li and Peiwen Yuan and Shaoxiong Feng and Boyuan Pan and Xinglin Wang and Bin Sun and Heda Wang and Kan Li},
  journal={ArXiv},
  year={2024},
  volume={abs/2401.10480},
  url={https://api.semanticscholar.org/CorpusID:267060971}
}

@inproceedings{sheng2025hybridflow,
  title={Hybridflow: A flexible and efficient rlhf framework},
  author={Sheng, Guangming and Zhang, Chi and Ye, Zilingfeng and Wu, Xibin and Zhang, Wang and Zhang, Ru and Peng, Yanghua and Lin, Haibin and Wu, Chuan},
  booktitle={Proceedings of the Twentieth European Conference on Computer Systems},
  pages={1279--1297},
  year={2025}
}

@article{Lee2025HowWD,
  title={How Well do LLMs Compress Their Own Chain-of-Thought? A Token Complexity Approach},
  author={Ayeong Lee and Ethan Che and Tianyi Peng},
  journal={ArXiv},
  year={2025},
  volume={abs/2503.01141},
  url={https://api.semanticscholar.org/CorpusID:276742093}
}

@article{Evans2003-EVAITM-3,
	author = {Jonathan St B. T. Evans},
	doi = {10.1016/j.tics.2003.08.012},
	journal = {Trends in Cognitive Sciences},
	number = {10},
	pages = {454--459},
	title = {In Two Minds: Dual-Process Accounts of Reasoning},
	volume = {7},
	year = {2003}
}

@article{feng2012dynamic,
  title={Dynamic programming for structured continuous Markov decision problems},
  author={Feng, Zhengzhu and Dearden, Richard and Meuleau, Nicolas and Washington, Richard},
  journal={arXiv preprint arXiv:1207.4115},
  year={2012}
}

@article{Grattafiori2024TheL3,
  title={The Llama 3 Herd of Models},
  author={Aaron Grattafiori and Abhimanyu Dubey and Abhinav Jauhri and Abhinav Pandey and Abhishek Kadian and Ahmad Al-Dahle and Aieleen Lakber and Aishwarya Selvaraj and Aitana De Las Cuevas Martinez and others},
  journal={ArXiv},
  year={2024},
  volume={abs/2407.21783},
  url={https://api.semanticscholar.org/CorpusID:271571434}
}
